\documentclass[journal]{IEEEtran}
\usepackage{blindtext}
\usepackage{graphicx}
\usepackage{xcolor}
\usepackage{colortbl}
\usepackage{rotating}
\usepackage{multirow}
\usepackage{booktabs}
\usepackage{bigstrut}
\usepackage{subcaption}
\usepackage{hyperref}
\usepackage{amsmath}
\usepackage{lscape}

\ifCLASSINFOpdf
\else
\fi
\begin{document}
%
\title{Multi-label Pixelwise Classification for Reconstruction of Large-scale Urban Areas}
%
%
%
\author{Yuanlie~He,~\IEEEmembership{Member,~IEEE,}
        Sudhir~Mudur,~\IEEEmembership{Sr. Member,~IEEE,}
        and~Charalambos~Poullis,~\IEEEmembership{Member,~IEEE}
\thanks{S. Mudur, C. Poullis are with the Department of Computer Science and Software Engineering, Concordia University, Quebec H3G 1M8 , Canada}
\thanks{Y. He is with the Guangdong University of Technology, University-Town, Guangzhou, P. R. China (510006)
}
\thanks{Manuscript received June 7, 2017; 
}}

%
%

\markboth{Journal of \LaTeX\ Class Files,~Vol.~6, No.~1, January~2007}%
{Shell \MakeLowercase{\textit{et al.}}: Bare Demo of IEEEtran.cls for Journals}
%



\maketitle

\begin{abstract}
Object classification is one of the many holy grails in computer vision and as such has resulted in a very large number of algorithms being proposed already. Specifically in recent years there has been considerable progress in this area primarily due to the increased efficiency and accessibility of deep learning techniques. In fact, for single-label object classification [i.e. only one object present in the image]  the state-of-the-art techniques employ deep neural networks and are reporting very close to human-like performance. 

There are specialized applications in which single-label object-level classification will not suffice; for example in cases where the image contains multiple intertwined objects of different labels. In this paper, we address the complex problem of multi-label pixelwise classification. We present our distinct solution based on a convolutional neural network (CNN) for performing multi-label pixelwise classification and its application to large-scale urban reconstruction. A supervised learning approach is followed for training a 13-layer CNN using both LiDAR and satellite images. An empirical study has been conducted to determine the hyperparameters which result in the optimal performance of the CNN. Scale invariance is introduced by training the network on five different scales of the input and labeled data. This results in six pixelwise classifications for each different scale. An SVM is then trained to map the six pixelwise classifications into a single-label. Lastly, we refine boundary pixel labels using graph-cuts for maximum a-posteriori (MAP) estimation with Markov Random Field
(MRF) priors. The resulting pixelwise classification is then used to accurately extract and reconstruct the buildings in large-scale urban areas. The proposed approach has been extensively tested and the results are reported. 
\end{abstract}


%
\IEEEpeerreviewmaketitle

\section{Introduction}
\label{sec:introduction}
Recent advances in the efficiency and accessibility of deep learning techniques have had a significant impact on the progress of many important and at the time dormant problems in computer vision. In particular, object recognition has greatly benefited since for many years the state-of-the-art had been almost restricted to minimal and incremental progress whereas currently human-like performances in object recognition \cite{ouyang2016factors, ren2015faster} are being reported, albeit for images with a single object.

Many successful applications which rely on single-label object recognition using deep neural networks have already been reported. The assumption with single-label object recognition is that a network can be trained to recognize objects from various categories and identify their general location [in the form of a bounding box] provided that the image contains exactly one object. More recently, it has been shown how this learning pipeline can be extended to handle cases where multiple spatially separable objects are present in an image i.e multi-label object recognition \cite{yangexploit}. However, in certain applications the images may contain objects which are intertwined. Moreover, rather than the general location, the precise location of the object is required. One such application is the classification of geospatial objects for reconstruction of large-scale urban areas. The data is in the form of geometry captured with LiDAR and satellite RGB images. The geospatial objects present in the data are buildings, roads, trees, artificial ground, natural ground, cars, etc and all these are perfectly intertwined i.e. a building is surrounded by ground, etc. One can think of the data (LiDAR, images) as being perfectly tesselated by many objects from each of these categories.   

In this paper we address the problem of multi-label pixelwise classification for large scale urban reconstruction and propose our distinct solution based on a 13-layer convolutional neural network (CNN). The CNN is trained using both LiDAR and satellite RGB images in multi-scale format, capturing large-scale urban areas and produces an output of six pixelwise values which are interpreted as likelihoods of the pixel in being a building, road, tree, car, or ground [artificial, natural]. An SVM linear classifier takes the likelihoods as inputs and maps them to a single label. In the final step, boundary pixel labels are refined.

Our technical contributions are:
\begin{itemize}
	\item the design, development and supervized training of a 13-layer convolutional network. The CNN is specifically designed for the classification of geospatial objects appearing in remote sensor data and in particular LiDAR and satellite RGB images into the following six classes: buildings, roads, tree, cars, natural ground and artificial ground.
	\item a method for introducing scale invariance during the training. Due to the multiple scales the CNN produces a likelihood-per-scale per pixel. These are further processed and combined into a single label by training an SVM. The labels are finally refined using graph cuts for maximum a-posteriori (MAP) estimation with Markov Random Field (MRF) priors which also addresses the problem of boundary pixels not being assigned labels.
	\item a complete framework for the geospatial object classification and reconstruction of large-scale urban areas. The multi-label pixelwise classification is used to reconstruct the buildings of large-scale urban areas. Generic objects such as cars and trees are replaced by procedurally generated models to yield realistic 3D visualizations.
\end{itemize}

\subsection{Paper Organization}
\label{subsec:paper_organization}
The paper is organized as follows: Section \ref{sec:related_work} presents an overview of the state-of-the-art in the area of object recognition and large-scale urban reconstruction. In Section \ref{sec:technical_overview} we present a technical overview of our proposed technique and in Section \ref{sec:dataset} we provide a brief description of the dataset used. The architecture of the developed network is described in detail in Section \ref{sec:cnn_architecture} including the training, refinement, validation and classification results. Section \ref{sec:urban_reconstruction} presents how these classification results are used in the context of large-scale urban reconstruction. The conclusion and future work are discussed in Section \ref{sec:conclusion}.

\section{Related Work}
\label{sec:related_work}
Object classification has been an active research topic in computer vision for many years and large-scale urban reconstruction for even more. In fact, object recognition is often employed as the first step in reconstruction for identifying the geospatial objects present in the scene. In this section we provide a brief overview of the state-of-the-art in the areas related to this work in object recognition using neural networks and in large-scale urban reconstruction. 

\subsection{Object Classification}
\label{subsec:object_classification}
The first Convolutional Neural Networks (CNN) was introduced by LeCun \cite{NeuralComput:LeCun} for hand writing recognition and could achieve outstanding performance. Yang et al. \cite{IEEE:Y.Yang} later extented the CNN with an additional layer for Support Vector Machine (SVM) which could detect and classify traffic signs. They demonstrated excellent performance and reported classification accuracies of $98.24\%$ and $98.77\%$ for the GTSDB and CTSD datasets, respectively. In a different application of object classification, Ijjina and Chalavadi \cite{Elsvier:E.P.Ijjina} proposed a method for recognizing human action. Instead of random initialization of the network they propose computing the initial weights of the CNN using genetic algorithms which minimize the classification error; using this method they can achieve classification accuracies of $99.98\%$and $96.92\%$ for the UCF50 and MNIST datasets. Hu et al. \cite{Elsvier:Y.Hu} propose a Single Signal Crowd CNN Model for counting dense crowds, and report outstanding performance for the training on the UCSD dataset and testing on the UCF-CROWD dataset. Liang et al. \cite{IEEE:M.Liang} propose a recurrent CNN (R-CNN) for object recognition by incorporating recurrent connections into each convolutional layer. The R-CNN is shown to outperform the state-of-the-art models on the CIFAR-10, CIFAR-100, MNIST and SVHN datasets.

More recently \cite{FCN} Fully-Convolutional Networks have been shown to produce the best results for multi-label pixel-wise classification by training on overlapping patches, however a significant disadvantage is the fact that the produced output is considerably downsampled compare to the input and further processing is required. When dealing with semantic segmentation of fine structures such as the ones appearing in urban datasets this generates spurious results. Of similar performance but same shortcoming is the CRF-based approach proposed in \cite{CRF} where again upsampling/interpolation is required on the generated output.

\subsection{3D reconstruction}
\label{subsec:3d_reconstruction}
The state-of-the-art in urban reconstruction can be better categorized according to the type and scale of the input data. For a comprehensive survey of urban reconstruction of various types and scales we refer the reader to the survey by Musialski et al \cite{Wiley:P.Musialski}. In this section, we provide a brief overview of state-of-the-art in large-scale urban reconstruction from remote sensor data, most relevant to our work.  

In \cite{zhou2012}, Zhou et al propose an automated system which given the exact bounding volume of a building can simplify the geometry based on dual contouring while retaining important features. Using this technique the authors were able to simplify the original geometry considerably. A different technique was presented in \cite{poullis2009} where pointcloud data was converted automatically to polygonal 3D models. This technique was applicable directly on the raw pointcloud data without requiring any user interaction. Later, in \cite{poullis2013} the authors extended the work to include a fast boundary refinement algorithm based on graph-cuts which was used to refine the boundaries. 

On a similar line of research, Lafarge et al. \cite{IEEE:L.AfargeF.} proposed a method which produces excellent reconstructed models from aerial LiDAR which can also handle the vegetation and complex grounds. Following a more interactive approach, Arikan et al  \cite{EUROGRAPHICS:A.LsisanS.} proposed a system for generating polyhedral models from semi-dense unstructured point-clouds. Planar surfaces were first extracted automatically based on prior semantic information, and later refined manually by an operator. 

The Achilles' heel of almost all reported work in this area is the geospatial object classification: if an object is misclassified then subsequent steps will most definitely also go wrong. Furthermore, extracting buildings from LiDAR data often produces jagged boundaries which affects the accuracy and quality of the reconstruction. Hence, it is of imperative importance to have as accurate classification as possible at the pixel-level. The proposed neural network achieves this yielding average accuracy in the high ninety percentile for buildings.   


\section{System Overview}
\label{sec:technical_overview}
The training dataset is first converted to the input form expected by the network. Scale invariance is introduced by training the network on  composite images containing multiple scales of the original depth map and RGB image captured by airborne LiDAR and satellite imaging. For training data, this also involves creation of multi-scale label data. 

The overall size of each of these composite images is very large. Hence random samples of a fixed patch size taken from the composite images are used for training the CNN. The CNN's output is six floating-point values per input pixel. These are interpreted as the likelihoods of the pixel to being classified with one of the six labels: building, road, tree, car, artificial ground, or natural ground. These likelihoods are used to train a linear classifier which outputs a single label per pixel. In a final step boundary labels are recovered and all labels are refined using graph-cuts for maximum a-posteriori (MAP) estimation with Markov Random Field
(MRF) priors.

The training of the network (CNN, linear classifier) is performed on a large urban dataset described in the following Section \ref{sec:dataset}. Once the network has been trained, data not used during the training is processed and labels are generated. The resulting labels are then used to extract only the buildings, cars and trees which are further processed to produce the 3D models representing the urban area.


\section{Dataset}
\label{sec:dataset}
\begin{figure}[!ht]
    \begin{subfigure}[t]{0.15\textwidth}
        \includegraphics[scale=0.2]{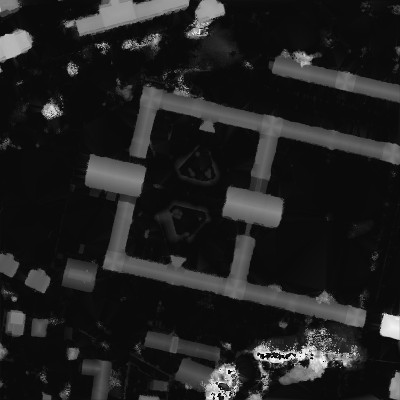}
        \caption{Depth map; contains values ranging from [0,1].}
        \label{fig:sample_depth_map}
    \end{subfigure}
    ~
    \begin{subfigure}[t]{0.15\textwidth}
        \includegraphics[scale=0.2]{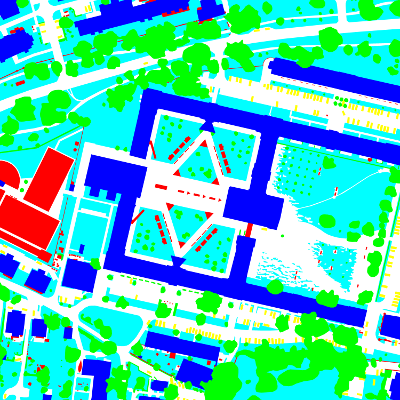}
        \caption{Label map; contains one of six values corresponding to the geospatial feature classes.}
        \label{fig:sample_label_map}
    \end{subfigure}
    ~
    \begin{subfigure}[t]{0.15\textwidth}
        \includegraphics[scale=0.2]{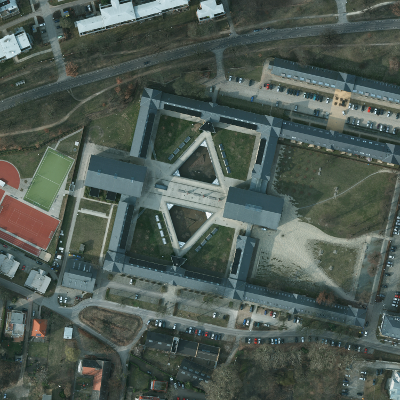}
        \caption{Color image; contains an RGB color where each channel ranges from [0,1].}
        \label{fig:sample_color_map}
    \end{subfigure}
    \caption{The Potsdam dataset consists of 24 image pairs. Each pair consists of a $6K\times6K$ (a) depth map, (b) label map and (c) color image.}
    \label{fig:data_sample}
\end{figure}

The data used for the training and testing of the proposed network is provided by the International Society for Photogrammetry and Remote Sensing (ISPRS). The data is available as part of the benchmark on urban object detection and 3D building reconstruction \cite{ISPRS:benchmark} and consists of several datasets. In this work, we have used the Potsdam dataset for 2D semantic labeling \cite{ISPRS:website} because of its higher accuracy. The Potsdam dataset consists of 24 image pairs consisting of three $6K\times6K$ registered images, namely a depth map captured with airborne LiDAR with a sampling density of $5cms$, a color satellite image, and the ground truth label map. The label map shows the ground-truth per-pixel classification into six classes: buildings (blue), trees (green), roads (white), natural ground (cyan), artificial ground (red) and cars (yellow). The 'artificial ground' label contains all areas on the ground that do not correspond to roads but are covered by materials such as asphalt that are typically used for paving roads. In particular, it contains parking lots, pavements, inner courtyards and driveways (if paved). The 'natural ground' label contains any areas on the ground covered by vegetation other than trees. In particular, it contains lawn and low bushes. The remaining labels are self-explanatory. Figure \ref{fig:data_sample} shows a sample pair available in the Potsdam dataset.

\section{Network Architecture}
\label{sec:cnn_architecture}
\begin{figure}[!ht]
    \vspace{-20pt}
    \includegraphics[scale=0.54525]{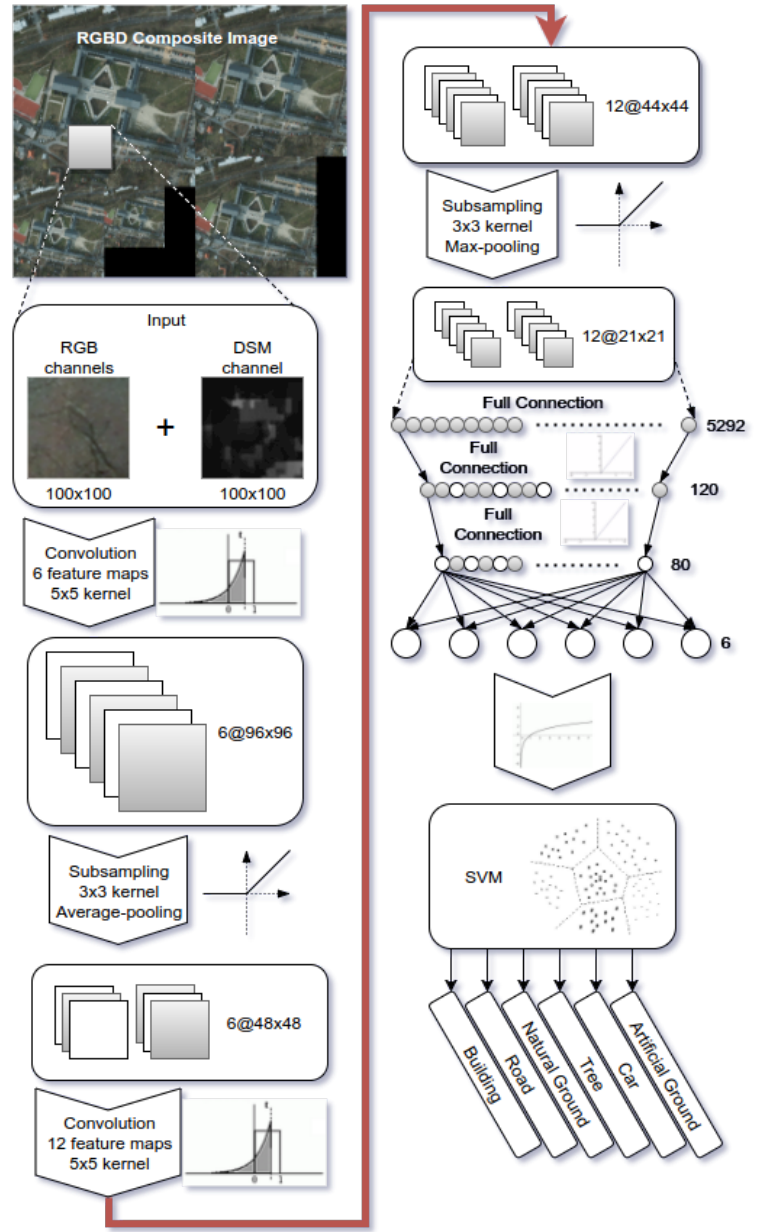}
    \caption{\label{fig:cnn_architecture} The proposed deep neural network for geospatial object classification of remote sensor data. A 13-layer CNN followed by a multi-class SVM can perform multi-label pixelwise classification into one of six labels: buildings, roads, trees, cars, ground [natural, artificial].}
\end{figure}

The proposed deep neural network consists of a 13-layer Convolutional Neural Network (CNN) and a linear classifier (SVM). A diagram of the network's architecture is shown in Figure \ref{fig:cnn_architecture}. 

\begin{figure}
    \centering
    \includegraphics[width=0.5\textwidth]{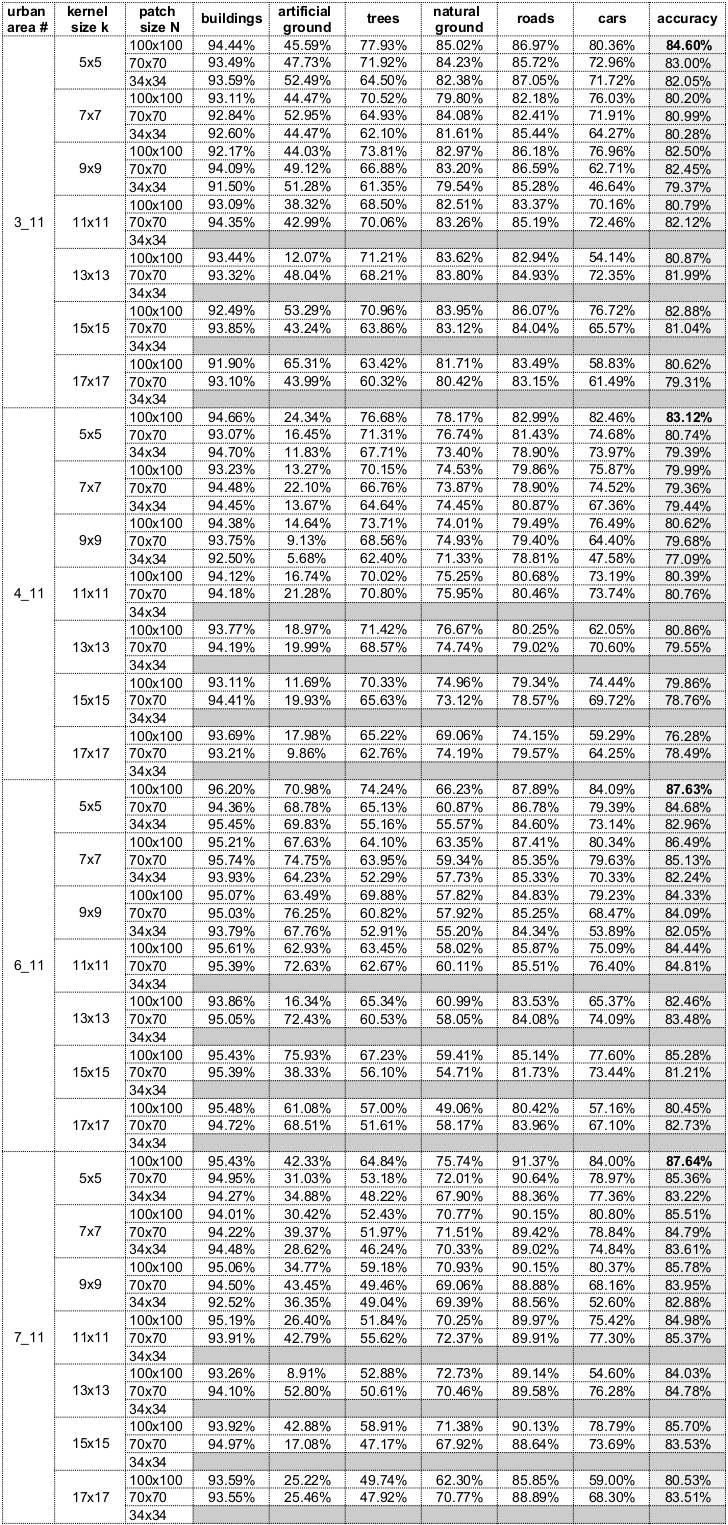}
    \caption{Gray cells: Kernel vs patch size is too small. The resulting image size after the two convolutional layers is too small and can not be processed further without changing the network architecture.}
    \label{fig:empirical_study}
\end{figure}

The input to the network are RGBD values corresponding to pixels contained in a small patch $N\times N$ of the input image. The optimal size of the patch $N\times N$ is determined empirically by varying the size while assessing the performance. Our experiments [the most relevant of which are shown in Figure \ref{fig:empirical_study}] have shown that the value resulting in optimal performance is $N=100$. The CNN's filter's kernel size $k$ was also determined in a similar fashion and is set to $k = 5$. An extensive empirical study showing the performance of the CNN with respect to different combinations of $3$ patch sizes [$34, 70, 100$] and $7$ kernel sizes [$5,7,9,11,13,15,17$] was performed. Table \ref{tab:empirical} shows the optimal performance achieved by the CNN for which the hyperparameters are set to patch size of $100\times 100$ and kernel size of $5 \times 5$.

\begin{table}
\centering
\hskip-0.7cm
\begin{tabular}{|p{10pt}|p{10pt}|p{20pt}|p{20pt}|p{20pt}|p{20pt}|p{20pt}|p{20pt}|p{20pt}|}
\hline
Urban Area & Ref. & Bldgs (\%) & Artif. Gnd.(\%) & Trees(\%) & Nat. Gnd.(\%) & Roads(\%)  & Cars(\%) & Acc. (\%)\\ 
\hline
P3 & w/o & 93.32 & 48.04 & 68.21 & 83.80 & 84.93 & 72.35 & 81.99\\ 
\hline
 & w & 93.44 & 48.20 & 68.79 & 84.16 & 85.23 & 74.14 & 82.34\\ 
\hline
P4 & w/o & 94.19 & 19.99 & 68.57 & 74.74 & 79.02 & 70.60 & 79.55\\ 
\hline
 & w & 94.41 & 20.51 & 69.36 & 75.14 & 79.33 & 72.58 & 79.99\\ 
\hline
P6 & w/o & 95.05 & 72.43 & 60.53 & 58.05 & 84.08 & 74.09 & 83.48\\ 
\hline
 & w & 95.30 & 73.16 & 61.67 & 58.41 & 84.34 & 75.94 & 83.88\\ 
\hline
P7 & w/o & 94.10 & 52.80 & 50.61 & 70.46 & 89.58 & 76.28 & 84.78\\ 
\hline
 & w & 94.26 & 54.40 & 50.82 & 70.65 & 89.73 & 77.73 & 85.01\\ 
\hline\end{tabular}
\label{table:table}
\caption{\small{\label{tab:empirical} Optimal network performance. The hyperparameters were empirically derived: patch size is $100\times 100$ and kernel size is $5 \times 5$. The table also shows a comparison between before(w/o) and after(w) applying the label refinement process.}} 
\end{table}

Based on the above, a patch of pixels with size $100\times100$ and four channels per pixel (RGBD) becomes the input to the network for all reported results.  Similarly a kernel size of $5\times5$ is used for all spatial convolutions. 

The first spatial convolution layer maps the input image patch into 6 feature maps [of size $96\times96$]. The following sub-sampling layer samples the output image of the previous layer with $3\times3$ kernel and average pooling, and generates 6 images [of size $48\times48$] output. The next layer of the network applies a spatial convolution and maps the 6 input images to 12 output images [of size $44\times44$]. Then these images are sub-sampled with 3$\times$3 kernel max pooling, and 12 ($21\times21$) mapping images are generated. All the pixels of the resulting 12 images are fully connected to a linear layer with 5292 nodes i.e. $12\times21\times21$. Next, the 5292-node linear layer passes through two fully connected linear layers of 120-nodes and 80-nodes respectively. The final linear layer is fully connected with the previous and consists of 6 nodes corresponding to the six labels. Each convolutional operation is followed by the non-linear operation $ReLU(x) = max(x,0)$. Thus, the CNN models the following operation,
\begin{equation}
    \Gamma (\Gamma (\Gamma (\Pi_{max} (ReLU(\Psi \star \Pi_{avg} (ReLU(\Psi \star X))))))) \mapsto \Phi^{p}
\end{equation}
where $p$ is a pixel and $\Phi^{p}$ is a 6-tuple of values corresponding to the six labels. In the above equation $X$ denotes the input data, $\Psi$ denotes a convolution kernel, $\Gamma(.)$ maps the input to a fully connected linear layer, $\Pi_{max}(.)$ is the max-pooling operation, $\Pi_{avg}(.)$ is the average-pooling operation, and $ReLU(.)$ is the rectified linear unit function.

As previously mentioned, the output of the CNN network is a 6-tuple $\Phi$ of values for each pixel $p$ contained in the input patch
\begin{equation}
    \Phi^{p} = <\phi^{p}_{1},...,\phi^{p}_{6}>
\end{equation}
where each component in $\Phi^{p}$ is interpreted as the \textit{unnormalized} likelihood of the pixel $p$ to be classified with any one of the six labels. $\phi^{p}_{1}...\phi^{p}_{6}$ represent the unnormalized probabilities of building, tree, road, artificial ground, natural ground and car respectively. We define $\Lambda^{p}$ as the 6-tuple of normalized likelihoods given by,
\begin{equation}
    \Lambda^{p} = \frac{e^{\Phi^{p}}}{\omega} = <\lambda^{p}_{1},...,\lambda^{p}_{6}> = <\frac{e^{\lambda^{p}_{1}}}{\omega},..., \frac{e^{\lambda^{p}_{6}}}{\omega}>
\end{equation}
where $\omega = \sum_{i=1}^{6} e^{\phi^{p}_{i}}$ such that $\sum_{i=0}^{6} \lambda^{p}_{i} = 1$.

\begin{figure*}[!ht]
    \begin{subfigure}[t]{0.33\textwidth}
        \includegraphics[scale=0.42]{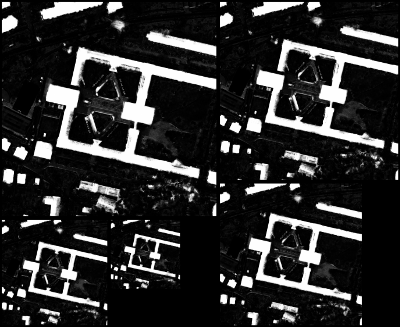}
        \caption{Buildings}
        \label{fig:likelihood_buildings}
    \end{subfigure}
    ~
    \begin{subfigure}[t]{0.33\textwidth}
        \includegraphics[scale=0.42]{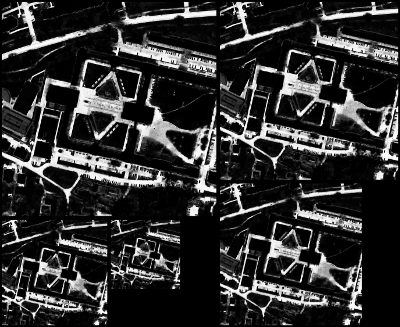}
        \caption{Roads}
        \label{fig:likelihood_roads}
    \end{subfigure}
    ~
    \begin{subfigure}[t]{0.33\textwidth}
        \includegraphics[scale=0.42]{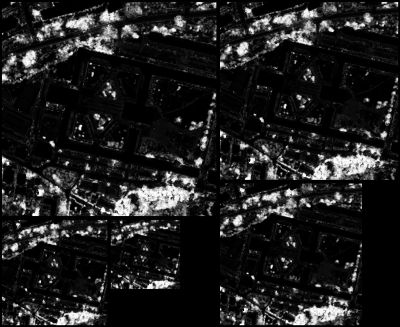}
        \caption{Trees}
        \label{fig:likelihood_trees}
    \end{subfigure}
    \newline
    \begin{subfigure}[t]{0.33\textwidth}
        \includegraphics[scale=0.42]{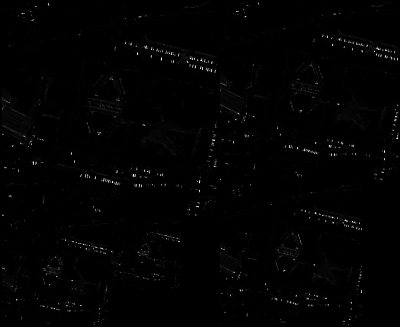}
        \caption{Cars}
        \label{fig:likelihood_cars}
    \end{subfigure}
    ~
    \begin{subfigure}[t]{0.33\textwidth}
        \includegraphics[scale=0.42]{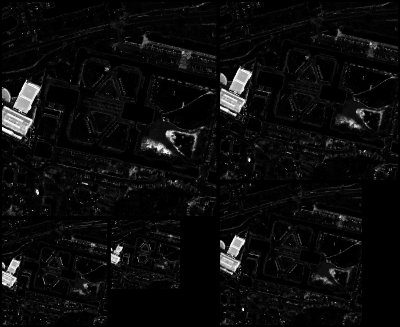}
        \caption{Artificial ground}
        \label{fig:likelihood_artificial_ground}
    \end{subfigure}
    ~
    \begin{subfigure}[t]{0.33\textwidth}
        \includegraphics[scale=0.42]{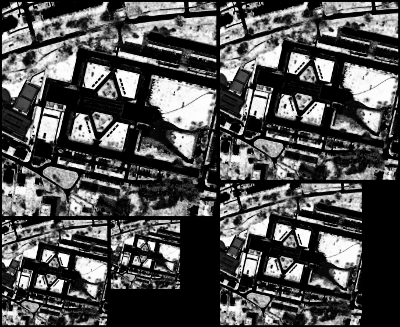}
        \caption{Natural ground}
        \label{fig:likelihood_natural_ground}
    \end{subfigure}
    \newline
    \begin{subfigure}[t]{0.33\textwidth}
        \includegraphics[scale=0.425]{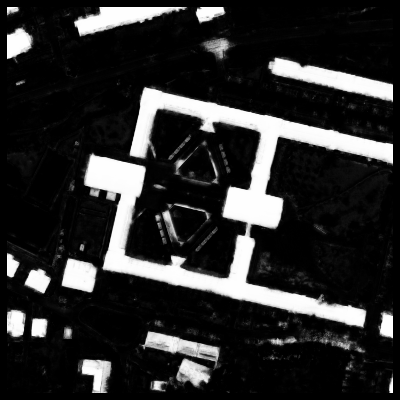}
        \caption{Combined likelihoods for label 'building'.}
        \label{fig:combined_likelihoods}
    \end{subfigure}
    ~
    \begin{subfigure}[t]{0.33\textwidth}
        \includegraphics[scale=0.425]{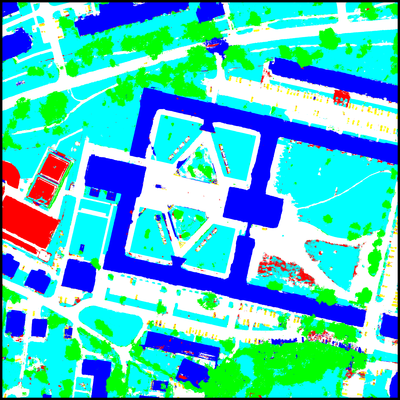}
        \caption{Resulting SVM label map.}
        \label{fig:generated_labels}
    \end{subfigure}
    ~
    \begin{subfigure}[t]{0.33\textwidth}
        \includegraphics[scale=0.425]{images/top_potsdam_3_11_label.png}
        \caption{The ground truth labels.}
        \label{fig:ground_truth_labels}
    \end{subfigure}
    \caption{(a)-(f): The per-pixel class likelihoods; intuitively, the brighter the value of a pixel in a class' image, the higher the likelihood of the pixel to be classified with that class. Bottom row: (g) The combined likelihoods for label 'building' which is used as part of the input to the SVM, (h) the per-pixel label classification map resulting from SVM; note that boundary pixels are not assigned a label at this point, (i) the ground truth labels.}
    \label{fig:likelihoods}
\end{figure*}

An example of the output is shown in Figure \ref{fig:likelihoods}. The components of the per-pixel likelihoods $\Lambda^{p}$ are grouped according to the labels and are shown as six images. The range of values of each individual component of $\Lambda$ is [0,1]. 

As it is evident from Figure \ref{fig:likelihoods}, the output at this point is a tuple $\Lambda$ for each pixel in the input \textit{composite} image. This means that for each pixel in the original [non-composite] image there will be essentially five sets of likelihoods within each composite image; one for each scale. The five sets of likelihoods $\Lambda_{i}$ with $1 \leq i \leq 5$ corresponding to each pixel are combined into a single tuple $\bar{\Lambda}$ by first up-scaling the images to the original $6k \times 6k$ resolution and then averaging the per-pixel likelihoods. Pixels lying on the boundaries for which not every scale may output a likelihood are not assigned likelihoods. 

The resulting  $6k \times 6k$ set of normalized likelihoods $\bar{\Lambda}^{p}$ corresponding to each pixel $p$ and the original label map with the same resolution $6k \times 6k$ becomes the input to a linear classifier (SVM). After training, the SVM learns weights $W$ and bias $b$ of the mapping function $f(W \times \bar{\Lambda}^{p} + b) \mapsto l_{i}$ where  $l_{i}, 1 \leq i \leq 6$ indicates one of the six labels. Figure \ref{fig:combined_likelihoods} shows the result of this process on the likelihoods corresponding to the label 'building'. Figure \ref{fig:generated_labels} shows the final output of the SVM and Figure \ref{fig:ground_truth_labels} shows the ground truth for the labels. It should be noted that at this point, boundary pixels cannot be assigned a label.

\subsection{Training}
\label{subsec:training}
The Potsdam dataset consists of 24 pairs of images. The training is performed on 20 randomly selected image pairs and validated against the remaining 4 image pairs. Inspired by DenseNet \cite{DenseNet}, we incorporate scale invariance into the training by preprocessing the original data to create composite images containing multiple resolutions of the original. These composite image pairs [depth, RGB, labels] become the input to the network. 
A composite contains the five images $I_{i}$ where $0\leq i\leq4$ each with resolution corresponding to $i\times16\%$ decrements of the original resolution i.e. $6k\times 6k, 5k\times 5k, 4k\times 4k, 3k\times 3k, 2k\times 2k$. The dataset contains considerable variance in the orientations of the geospatial objects hence rotation invariance is implicitly incorporated in the training.

\begin{figure}[!ht]
    \centering
    \includegraphics[scale=0.6]{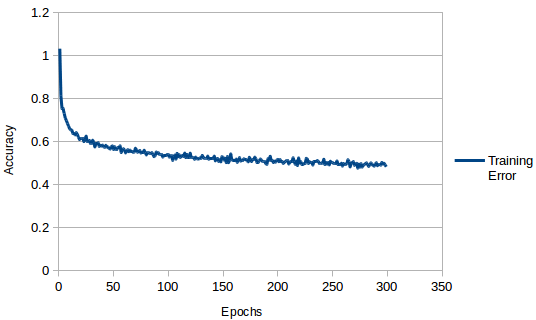}
    \caption{Training error of 300 iterations}
    \label{fig:training}
\end{figure}

The CNN was trained for 300 epochs on a single machine with the following specifications: Intel Core i7-6700K CPU @ 4.00GHz × 8, 16GB RAM, 12GB NVidia GeForce GTX TITAN X/PCIe/SSE2. The Torch API was used for the development of the CNN and the code will be made available as open source. The duration of the training for 300 epochs was 26 hours however, as it can be seen from Figure \ref{fig:training} after the first few epochs the training error rapidly reduces and almost converges.

Although the available memory on the GPU is 12GB, the Torch API imposes a restriction on the maximum GPU usage to 2GB. Hence, the available training data cannot be used in its entirety. Instead, given the 20 training image pairs we perform random sampling and gather as many training samples [$100\times 100$ image patches] as the memory can fit. The uniform random sampling from the 20 image pairs includes patches from various resolutions. We ensure that all pixels within each sampled image patch fall entirely within the same scale. Sampled patches falling on boundaries between different scales are rejected. This results in a total of $400,000$ training samples.

Finally, the likelihoods generated by the CNN are combined as previously described and fed into multi-class SVM, which generates the pixel's classification after using a one-vs-all learning strategy. The SVM learns how to map the 6-tuple $\bar{\Lambda}^{p}$ corresponding to each pixel $p$ into a single class which classifies the input with the highest margin. The Torch API was again used for the development of the SVM and the code will be made available as open source. The duration of the training was $27$ minutes and was performed on the same machine as above.





\subsection{Maximum-a-posteriori Inference with Markov Random Field Priors for Label Refinement}
The linear classifier combines the six pixel-wise likelihoods produced by the CNN into a single label. Pixels along the boundaries of the image cannot be assigned a label because the convolutional filter falls out of bounds. To overcome this problem Overfeat \cite{overfeat} first introduced the shift-and-stitch trick where shifted versions of the input were processed and the results interleaved into a full resolution output. However, the computational efficiency of this approach does not scale to the current large-scale urban datasets we are dealing with. 

Instead, we propose the use of graph-cuts for maximum-a-posteriori (MAP) estimation with Markov Random Field (MRF) priors. We reformulate the problem as finding an optimal labeling $f_{p}$ for every pixel $p$ such that $f(p) \mapsto l$ where $l$ is a new label. In addition to the six labels we include a new label \textit{unknown} to account for the boundary pixels which have not been assigned a label. Hence, the set of labels becomes \textit{[buildings, roads, trees, cars, natural ground, artificial ground, unknown]}.

The energy function which is minimized is then given by,
\begin{equation}
    E(f) = E_{unary}(f) + E_{pairwise}(f)
\end{equation}

The unary energy term $E_{unary}(f)$ provides a measure of the compatibility of the new label under the labeling $f(p_{i})$ to the pixel $p_{i}$ with label $l_{p_{i}}$ in the observed data and is given by,
\begin{equation}
    E_{unary}(f) = \sum_{i=0}^{N}     
    \begin{cases}
        10, & \text{if $f(p_{i}) \neq l_{p_{i}}$}.\\
        15, & \text{if $f(p_{i}) = unknown$}.\\
         0, & \text{if $f(p_{i}) = l_{p_{i}}$}
    \end{cases}
\end{equation}

The pairwise energy term $E_{pairwise}(f)$ provides a measure of compatibility of the new labels under the labeling $f(p_{i}), f(p_{j})$ for neighbouring pixels $p_{i}$ and $p_{j}$ respectively and is given by,
\begin{equation}
    E_{pairwise}(f) = \sum_{i,j=0}^{N} 
    \begin{cases}
        0, & \text{if $l_{p_{i}} = l_{p_{j}}$}.\\
        20, & \text{otherwise}.\\
    \end{cases}
\end{equation}

\begin{figure}[!ht]
    \begin{subfigure}[t]{0.225\textwidth}
        ics[scale=0.3]{images/labels2.png}
        \caption{}
    \end{subfigure}
    ~
    \begin{subfigure}[t]{0.225\textwidth}
        \includegraphics[scale=0.3]{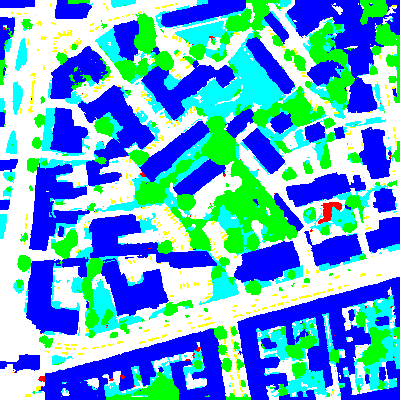}
        \caption{}
    \end{subfigure}
    \caption{Maximum-a-posteriori Inference with Markov Random Field Priors for Label Refinement\label{fig:graph-cuts}. (a) The labels generated by the network. Pixels along the image boundaries cannot be assigned a label. (b) Dense and refined labeling resulting from the proposed method. Pixels along the image boundary are assigned a label.}
\end{figure}

The pairwise measure $V(f_{p_{i}}, f_{p_{j}})$ between neighbouring pixels $p_{i}$, $p_{j}$, and $p_{k}$ can be trivially shown to be metric since the following conditions are true,
\begin{equation}
    \begin{split}
        V(f(p_{i}), f(p_{j})) = 0 \leftrightarrow i = j\\
        V(f(p_{i}), f(p_{j})) = V(f(p_{j}), f(p_{i})) > 0\\
        V(f(p_{i}), f(p_{j})) \leq V(f(p_{i}), f(p_{k})) + V(f(p_{k}), f(p_{j}))
    \end{split}
\end{equation}
This is a multi-label MRF problem with non-submodular energy potentials and as such can only be approximately solved. The alpha-expansion algorithm is used to break the multi-label problem into a series of binary problems. Experiments have shown that after 5 iterations the energy $E(f)$ is reduced on average $~12\%$ and the overall accuracy of the classification results increases by $[0.5-1\%]$. Table \ref{tab:empirical} shows a comparison between the performance before(w/o) and after(w) this process. Image boundary pixels for which no label was generated are now relabeled based on the new labeling resulting from the energy minimization using graph-cuts. Figure \ref{fig:graph-cuts} shows an example output of this process.

\vspace{-10pt}
\subsection{Validation}
\label{subsec:validation}

The proposed network was validated against two sets of test images: (a) the 4 image pairs out of the 24 available which were not used in the training and (b) the 14 image pairs available for testing for which ISPRS did not make the ground truth publicly available. The performance is measured in terms of Precision ($P$), Recall ($R$), and $F_{1}$ score which are defined as,
\begin{equation}
    \begin{split}
        P = \frac{tp}{tp+fp} \qquad
        R = \frac{tp}{tp+fn} \qquad
        F_{1} = 2 \times \frac{P \times R}{P + R}
    \end{split}
\end{equation}
where $tp$ indicates the true positives, $fp$ indicates the false positives, and $fn$ indicates the false negatives.

\subsubsection{Classification Results for 4 image pairs}
Out of the 24 available image pairs, 20 were used for the training and the remaining 4 were used for validation. The network performance statistics for the 4 validation images were presented in Table \ref{tab:empirical}. These statistics were computed from the ground truth labels made available with the Potsdam dataset. 

\subsubsection{Classification Results for 14 image pairs}
\begin{figure}[!ht]
    \begin{subfigure}[t]{0.15\textwidth}
        \includegraphics[scale=0.14]{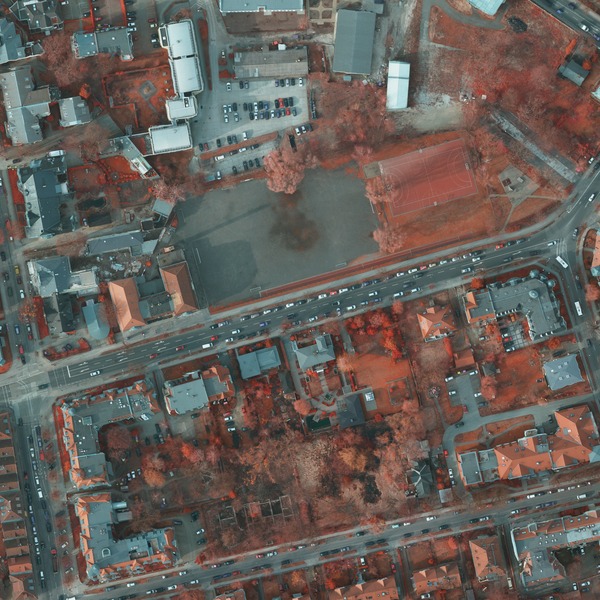}
        \caption{}
        \label{fig:sample4_14_satellite}
    \end{subfigure}
    ~
    \begin{subfigure}[t]{0.15\textwidth}
        \includegraphics[scale=0.14]{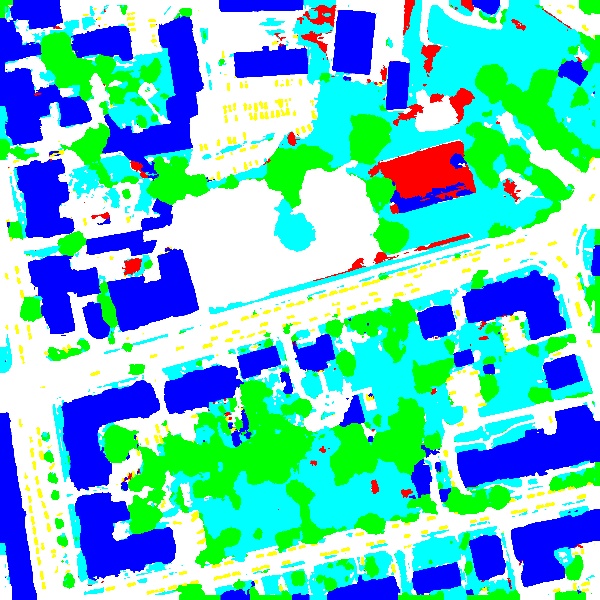}
        \caption{}
        \label{fig:sample4_14_labels}
    \end{subfigure}
    ~
    \begin{subfigure}[t]{0.15\textwidth}
        \includegraphics[scale=0.14]{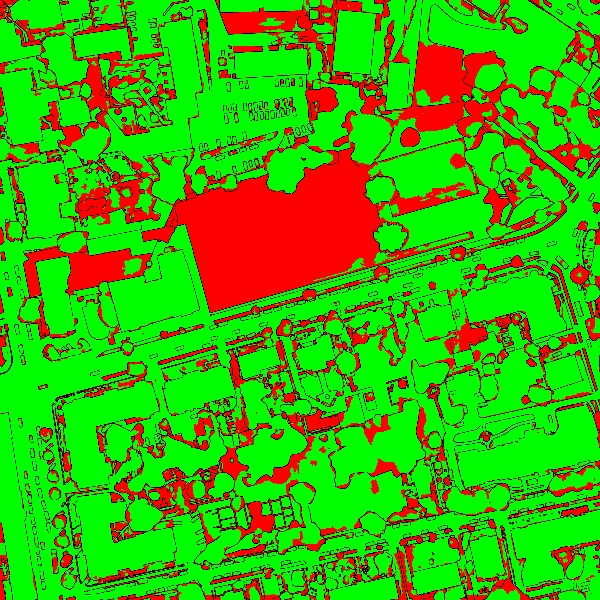}
        \caption{}
        \label{fig:sample4_14_red_green}
    \end{subfigure}
    \caption{\small{The evaluation result for one of the 14 test images. Resolution $6k\times6k$. (a) Satellite image of the urban area P4-14. (b) Generated label map. (c) Red/green image, indicating wrongly classified pixels.}}
    \label{fig:sample4_14}
\end{figure}

The ISPRS benchmark also contains 14 image pairs for which ground truth was not made publicly available. The following network performance statistics were computed by and provided by the ISPRS Working Group II/4 organizers as part of their urban classification benchmark. Table \ref{tab:test_results} shows the evaluation of the overall classification results for the 14 test images and as it can be seen the overall accuracy for building classification is almost $95\%$. Figure \ref{fig:sample4_14} shows the evaluation results for one of the 14 test images, namely P4-14. The individual evaluation results for P4-14 are shown in Table \ref{tab:sample4_14_evaluation}.

\begin{table}
\centering
\begin{tabular}{|p{40pt}|p{20pt}|p{20pt}|p{25pt}|p{20pt}|p{20pt}|p{20pt}|}
\hline
pred./ref. & roads & bldgs & nat. gnd. & tree & car & artif. gnd.\\ 
\hline
roads & 89.9 & 1.6 & 4.9 & 2.9 & 0.1 & 0.6\\ 
\hline
bldgs & 2.5 & 94.40 & 0.8 & 2 & 0 & 0.3\\ 
\hline
nat. gnd & 6.8 & 0.90 & 82.40 & 9.1 & 0 & 0.7\\ 
\hline
tree & 7.6 & 1 & 20.3 & 70.6 & 0.4 & 0.1\\ 
\hline
car & 17.60 & 1.6 & 0.8 & 1.80 & 76.7 & 1.7\\ 
\hline
artif. gnd. & 38.7 & 8.3 & 28.60 & 4.8 & 2 & 17.60\\ 
\hline
Prec./Corr. & 84.40 & 95 & 72.5 & 77.4 & 86.3 & 60.5\\ 
\hline
Rec./Compl. & 89.9 & 94.40 & 82.40 & 70.6 & 76.7 & 17.60\\ 
\hline
F1 & 87.1 & 94.70 & 77.10 & 73.9 & 81.2 & 27.3\\ 
\hline\end{tabular}
\caption{\label{tab:test_results} \small{The overall evaluation of the classification results for the 14 test images for which ground truth was not provided. The network performance statistics were computed by and provided by the ISPRS Working Group II/4 organizers as part of their urban classification benchmark. All shown values are percentages.}}
\end{table}

 \begin{table}
\centering
\begin{tabular}{|p{40pt}|p{20pt}|p{20pt}|p{25pt}|p{20pt}|p{20pt}|p{20pt}|}
\hline
pred./ref. & roads & bldgs & nat. gnd. & tree & car & artif. gnd.\\ 
\hline
roads & 89.44 & 1.17 & 6.04 & 1.82 & 0.11 & 1.42\\ 
\hline
bldgs & 1.80 & 96.61 & 0.38 & 0.87 & 0.09 & 0.24\\ 
\hline
nat. gnd. & 17.64 & 1.10 & 70.3 & 9.82 & 0.04 & 1.11\\ 
\hline
tree & 9.72 & 1.18 & 16.73 & 71.75 & 0.4 & 0.22\\ 
\hline
car & 18.51 & 1.7 & 0.48 & 1.69 & 76.02 & 1.59\\ 
\hline
artif. gnd. & 64.15 & 6.8 & 11.96 & 2.24 & 0.49 & 14.36\\ 
\hline
Prec./Corr. & 63.5 & 93.2 & 72.5 & 81.5 & 88.7 & 66.5\\ 
\hline
Rec./Compl. & 89.4 & 96.6 & 70.3 & 71.7 & 76 & 14.40\\ 
\hline
F1 & 74.2 & 94.90 & 71.40 & 76.3 & 81.90 & 23.60\\ 
\hline\end{tabular}
\caption{\label{tab:sample4_14_evaluation} \small{Evaluation results for urban area P4-14. All shown values are percentages.}} 
\end{table}

\section{Urban Reconstruction}
\label{sec:urban_reconstruction}
\begin{figure}[!ht]
    \begin{subfigure}[t]{0.45\textwidth}
        \includegraphics[scale=0.4]{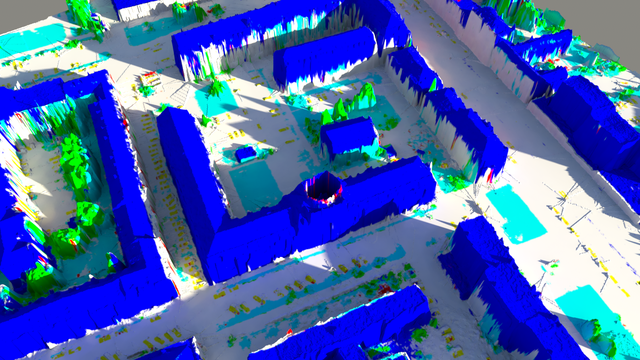}
        \caption{}
    \end{subfigure}
    
    \begin{subfigure}[t]{0.45\textwidth}
        \includegraphics[scale=0.4]{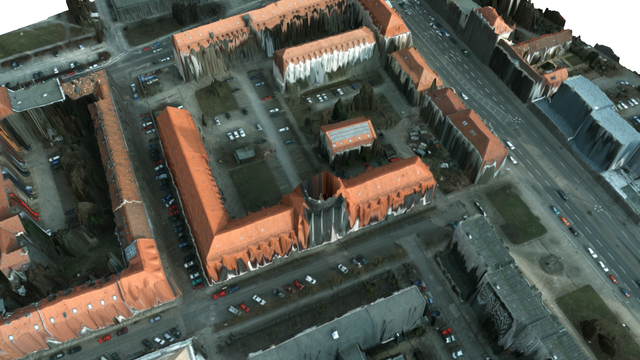}
        \caption{}
    \end{subfigure}
    
    \begin{subfigure}[t]{0.45\textwidth}
        \includegraphics[scale=0.4]{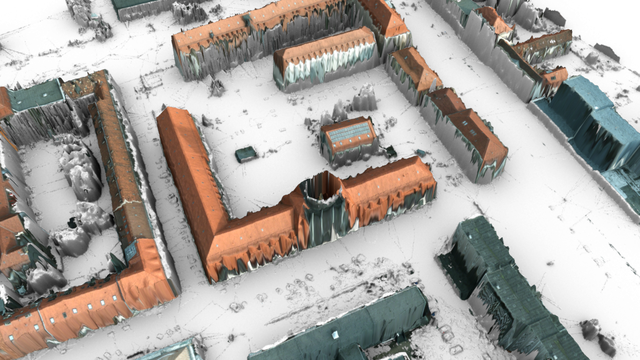}
        \caption{}
    \end{subfigure}
    \caption{(a) The resulting labels being projected onto the 3D models. (b) The satellite image projected onto the scene models. (c) Textured models showing only the classified buildings.}
    \label{fig:textured}
\end{figure}

\begin{figure}
        \centering
        \begin{subfigure}[t]{0.45\textwidth}
            \includegraphics[scale=0.35]{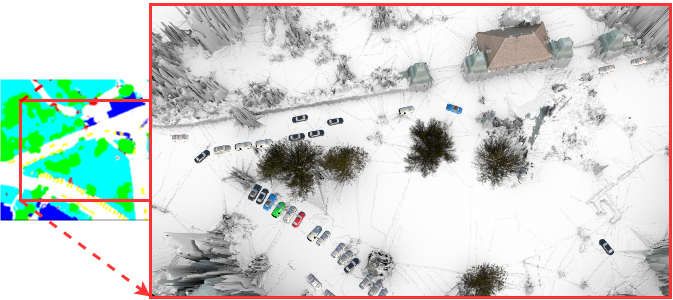}
            \caption{}
            \label{fig:closeup}
        \end{subfigure}
        \begin{subfigure}[t]{0.45\textwidth}
            \includegraphics[scale=0.38]{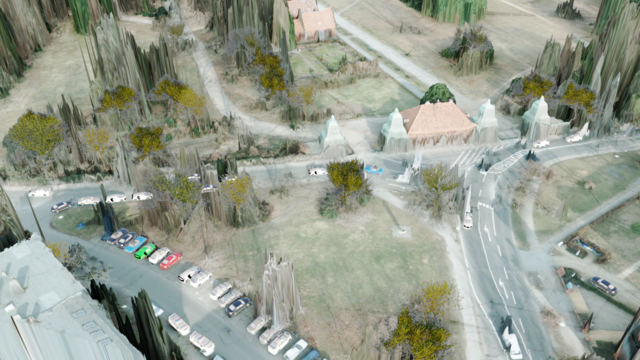}
            \caption{}
            \label{fig:render}
        \end{subfigure}
        \caption{Reconstructed and textured models. Buildings are replaced by polygonal models, cars by CAD models, and trees by procedural models. A mesh resulting from triangulating the depth map is used for the other classes: roads, natural and artificial ground.}
\end{figure}

The classification results are further processed. The depth map is used to extract boundary points and extrude 3D models to represent the geospatial objects in the scene. In particular the boundaries for the buildings are extracted and extruded to create polygonal models and generic objects such as cars are replaced by CAD models, and trees are replaced by procedural models. Figure \ref{fig:textured}(a)-(b) shows the generated labels being projected onto the same 3D models and Figure \ref{fig:textured}(c)-(d) shows textured models only for the classified buildings. Figure \ref{fig:closeup} shows a closeup of an urban area in which buildings have been replaced by polygonal models, cars by generic CAD models and trees by procedural models. The same scene with textures and from a different viewpoint is shown in Figure \ref{fig:render}.


\section{Conclusion}
\label{sec:conclusion}
We have presented a novel technique for multi-label pixel-wide classification for reconstruction of large-scale urban areas. Unlike existing methods, the proposed method relies on a relatively small CNN to efficiently process large sets of data. An empirical study was performed and presented to determine the parameters for which the network produces optimal results. Scale invariance is incorporated in the processing by training the network on composite images containing multiple scales of the originals. This results in multiple per-pixel classification labels which are mapped into a single label using a trained linear classifier. Pixels lying on image boundaries are not assigned any label. We reformulate the problem as a labeling problem and propose the use of graph-cuts for maximum-a-posteriori inference of those labels with Markov Random Field priors. The result is a dense set of labels where all pixels in the image are assigned a label according to the minimized energy function. The proposed technique has been extensively tested on large-scale datasets depicting urban areas for which ground truth is available. The achieved accuracy in the classification ranges in the $90^{th}$ percentile.


%
\section*{Acknowledgment}
The authors would like to thank the International Society for Photogrammetry and Remote Sensing and in particular the ISPRS Working Group II/4 organizers and Dr. Markus Gerke for creating the benchmark and making it publicly available. We also thank Microsoft Research for making processing large datasets possible through its Azure for Research Award.

\ifCLASSOPTIONcaptionsoff
  \newpage
\fi

%


\begin{IEEEbiography}[{\includegraphics[width=1in,height=1in,clip,keepaspectratio]{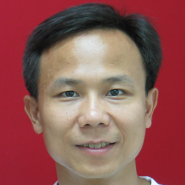}}]{Yuanlie He} was born in Guangdong, China, in 1976. He received the B.S. and M.S. degrees in electronic engineering from Wuyi University, China in 1998 and 2000, respectively. He recevied the Ph.D. degree in automation science and engineering from South China University of Technology, China in 2003.

He is an Associate Professor with the School of Computer, Guangdong University of Technology, China. He researched satellite image recognition in Concordia University, Canada as visited scientist in 2016. His research interests include computer vision, mobile robot and intelligent robot.
\end{IEEEbiography}

\begin{IEEEbiography}[{\includegraphics[width=1in,height=1in,clip,keepaspectratio]{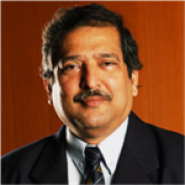}}]{Sudhir Mudur} obtained his Bachelor of Technology (honours) in 1970 from IIT Bombay and his PhD in 1976 from the Tata Institute of Fundamental Research in Mumbai, India. His interest in computer graphics started with his undergraduate thesis project. Since then he has been actively researching the field of computer graphics, particularly the areas 3D modelling, global illumination, virtual environments and applications in CAD/CAM and entertainment. Over this period of more than 4 decades, he has published papers in top computer graphics venues and supervised a large number of doctoral and graduate students, many of whom are well established in the field. His work in the areas of robust geometric computing using interval arithmetic and in global illumination models is well cited. Mudur is currently a professor and chair of the department of computer science and software engineering at Concordia university. Mudur is a senior member of IEEE, member of ACM, member of Eurographics and a SIGGRAPH Pioneer Group member.
\end{IEEEbiography}

\begin{IEEEbiography}[{\includegraphics[width=1in,height=1in,clip,keepaspectratio]{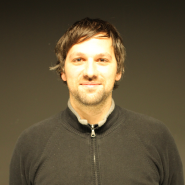}}]{Charalambos Poullis} was born in Nicosia, Cyprus, in 1978. He received the B.Sc. degree in Computing and Information Systems with First Class Honors from the University of Manchester, UK, in 2001, and the M.Sc. in Computing Science with specialisation in Multimedia and Creative Technologies, and Ph.D. in Computer Science from the University of Southern California (USC), Los Angeles, USA, in 2003 and 2008, respectively. Since August 2015, he has been with the Department of Computer Science and Software Engineering at the Faculty of Engineering and Computer Science at Concordia University where he also serves as the Director of the Immersive and Creative Technologies (ICT) lab, member of the 3D Graphics Group. His current research interests lie at the intersection of computer vision and computer graphics. More specifically, he is involved in fundamental and applied research covering the following areas: acquisition technologies \& 3D reconstruction, photo-realistic rendering, feature extraction \& classification, virtual \& augmented reality. Charalambos is a member of the Association for Computing Machinery(ACM); Institute of Electrical and Electronics Engineers (IEEE) Computer Society; Marie Curie Alumni Association (MCAA); ACM Cyprus Chapter, where he also served in the management committee between 2010-2015; and British Machine Vision Association (BMVA). Charalambos has been serving as a regular reviewer in numerous premier conferences and journals since 2003. 
\end{IEEEbiography}

\vfill


\end{document}